\documentclass[10pt,twocolumn,letterpaper]{article}
\usepackage[pagenumbers]{cvpr}
\usepackage{times}  
\usepackage{helvet}  
\usepackage{courier}  
\usepackage[hyphens]{url}  
\usepackage{graphicx} 
\usepackage{natbib}  
\usepackage{caption} 
\usepackage{algorithm}
\usepackage{algpseudocode}

\usepackage{newfloat}
\usepackage{listings}
\DeclareCaptionStyle{ruled}{labelfont=normalfont,labelsep=colon,strut=off}
\floatstyle{ruled}
\newfloat{listing}{tb}{lst}
\floatname{listing}{Listing}

\usepackage{amsmath, amssymb, amsfonts, mathtools}
\usepackage{bm} 
\usepackage{booktabs, multirow, makecell, colortbl}
\usepackage[table]{xcolor}
\definecolor{RowColor}{rgb}{0.97, 0.97, 1}
\definecolor{OursGreen}{RGB}{236,248,239}

\newif\ifshowseedstd
\showseedstdtrue

\newcommand{\resultcell}[2]{#1\,{\scriptsize$\pm$\,#2}}

\newcommand{\sdPeMSRMSE}{0.11}
\newcommand{\sdPeMSMAE}{0.09}
\newcommand{\sdPeMSMAPE}{0.12}

\newcommand{\sdHZRMSE}{0.15}
\newcommand{\sdHZMAE}{0.08}
\newcommand{\sdHZMAPE}{0.13}

\newcommand{\sdKnowRMSE}{0.08}
\newcommand{\sdKnowMAE}{0.05}
\newcommand{\sdKnowMAPE}{0.09}

\newcommand{\sdFLFiveRMSE}{0.18}
\newcommand{\sdFLFiveMAE}{0.12}
\newcommand{\sdFLFiveMAPE}{0.15}

\newcommand{\sdFLTenRMSE}{0.26}
\newcommand{\sdFLTenMAE}{0.18}
\newcommand{\sdFLTenMAPE}{0.22}

\usepackage{subcaption} 
\usepackage{enumitem}   
\usepackage{textcomp}
\usepackage{appendix}

\definecolor{cvprblue}{rgb}{0.21,0.49,0.74}
\usepackage[pagebackref,breaklinks,colorlinks,allcolors=cvprblue]{hyperref}
\hypersetup{
  pdftitle={F2STNet: Fair and Federated Spectral-Temporal Modeling for Graph Forecasting},
  pdfauthor={Jiayi Zhang, Jinfeng Xu, Hewei Wang, Siyuan Cen, Haidong Huang, Yiyao Zhan, Zheyu Chen, Jinjiang You, Ai Jian, and Edith C. H. Ngai}
}

\usepackage{pifont}

\title{F$^{2}$STNet: Fair and Federated Spectral-Temporal Modeling\\ for Graph Forecasting}
\author{
Jiayi Zhang$^{1,*}$ \enspace
Jinfeng Xu$^{2,*}$ \enspace
Hewei Wang$^{3,*}$ \enspace
Siyuan Cen$^{3}$ \enspace
Haidong Huang$^{1}$ \\
Yiyao Zhan$^{1}$ \enspace
Zheyu Chen$^{4}$ \enspace
Jinjiang You$^{3}$ \enspace
Ai Jian$^{1}$ \enspace
Edith C. H. Ngai$^{2}$ \\[0.5em]
{\small $^{1}$University of Nottingham \quad
$^{2}$The University of Hong Kong} \\
{\small $^{3}$Carnegie Mellon University \quad
$^{4}$The Hong Kong Polytechnic University}
}

\begin{document}

\maketitle

\begingroup
\renewcommand{\thefootnote}{\fnsymbol{footnote}}
\footnotetext[1]{Equal contribution.}
\endgroup

\begin{abstract}

Spatiotemporal prediction on graph-structured data is central to traffic forecasting and environmental monitoring, yet decentralized and heterogeneous data complicate both sequence modeling and collaborative training. We propose F$^2$STNet, a federated forecasting framework that combines truncated graph-Fourier features, a lightweight diagonal state-space temporal encoder, graph convolution, and Fairness-aware Federated Aggregation (FFA). The spectral branch exposes graph-frequency structure, while the state-space layer models long temporal dependencies with linear complexity in the sequence length. FFA adjusts the FedAvg prior using client validation losses and an increasing fairness schedule. Experiments on PeMS04, HZMetro, and KnowAir show favorable forecasting accuracy relative to the evaluated baselines; federated experiments on PeMS04 additionally improve worst-client and client-dispersion metrics.

\end{abstract}

\section{Introduction}

Spatiotemporal forecasting over graph-structured data plays a crucial role in traffic management, environmental monitoring, and urban planning. It requires jointly modeling spatial dependencies among nodes and temporal dynamics over time. Graph Neural Networks (GNNs), particularly Graph Convolutional Networks (GCNs), are widely adopted due to their strong ability to capture spatial relations~\citep{zhou2020graph}. While recent advances have proposed unified spatial-temporal architectures~\citep{tang2023modeling,ijgi12030100}, most methods still rely on centralized data access and resource-heavy recurrent or attention-based temporal modules, limiting their applicability in privacy-sensitive and resource-constrained environments. Beyond traffic and environmental sensing, temporally ordered observations also underpin precise video-language supervision and camera-motion understanding~\citep{Lin_2026_CVPR,lin2025towards}; multi-frame camera-array calibration further illustrates how measurements may be coupled across both viewpoints and time~\citep{you2025multi}.

Federated Learning (FL) offers a decentralized training paradigm that enables multiple clients to collaboratively learn a global model without sharing raw data. This setup is especially appealing for spatiotemporal tasks involving data silos, such as distributed traffic sensors or environmental monitors. However, applying FL to spatiotemporal forecasting introduces two major challenges: (i) statistical heterogeneity across clients degrades generalization and leads to biased global models, and (ii) uniform aggregation methods such as FedAvg~\citep{fedavg} fail to account for disparities in client difficulty, causing unfair performance degradation for underrepresented clients~\citep{liu2024fuels,fl_fairness_survey}. Related evidence from foundation-model transfer shows that noisy supervision can reshape learned feature spaces and consistently harm out-of-domain performance~\citep{10934976}, underscoring the need for robust learning under distributional variation.

\begin{figure}
    \centering
    \includegraphics[width=\linewidth]{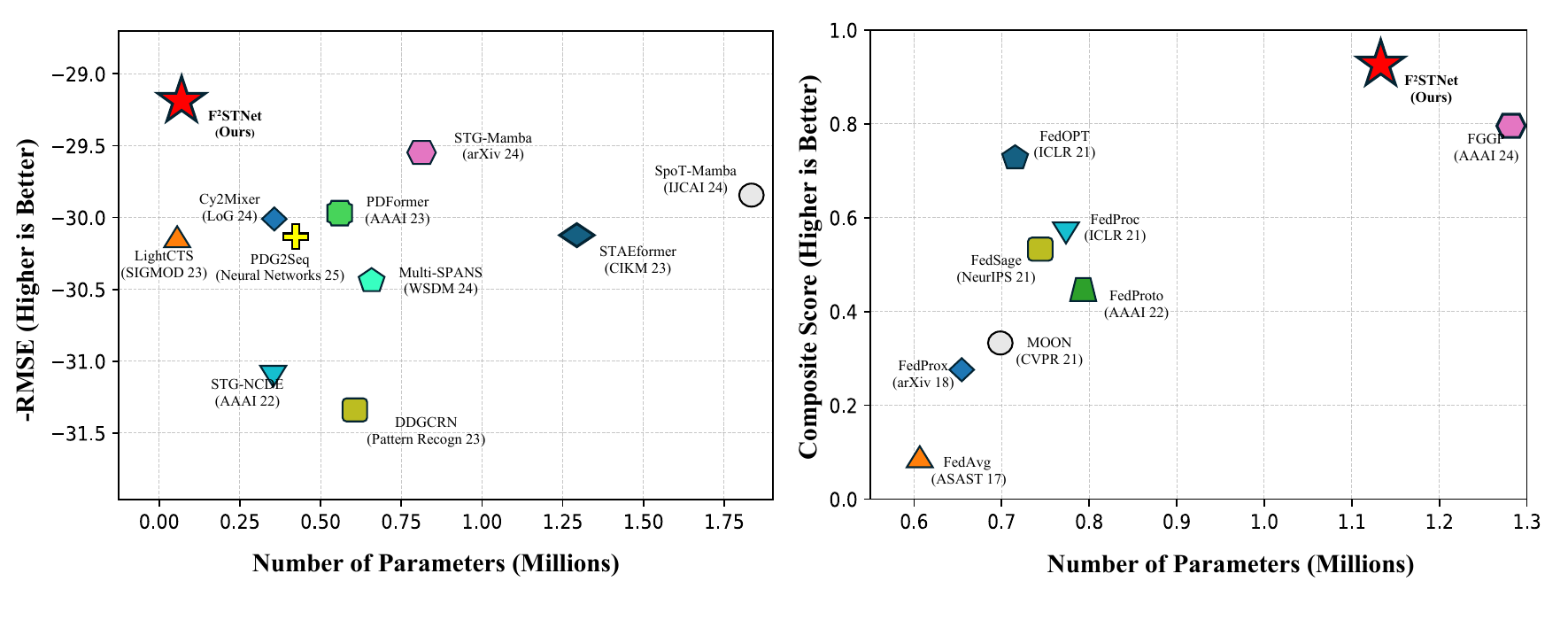}
    \caption{
    Performance–efficiency trade-off of spatiotemporal models (left) and FL methods (right).
    Left: RMSE vs. model size on PeMS04 shows accuracy–complexity trade-off; Right: score combining fairness, accuracy, and generalization.
    F$^2$STNet delivers top performance with low model size and cost, supporting deployment in centralized and federated settings under resource and fairness constraints.
    }
    \label{fig:leaderboard}
    \vspace{-5mm}
\end{figure}

While prior works have addressed individual challenges—such as spectral modeling~\citep{Cao2020StemGNN}, temporal dependencies~\citep{cy2mixer}, or fairness-aware aggregation~\citep{Ezzeldin2023FairFed}—these solutions remain fragmented. In particular, integrating them into a unified federated spatiotemporal framework combining frequency encoding, lightweight temporal modeling, and fairness-aware aggregation remains largely unexplored. This highlights the need for a holistic architecture capable of jointly addressing these challenges under federated constraints while ensuring scalability and robustness in deployment.

To address these challenges, we propose \textbf{F$^2$STNet}, a federated spatiotemporal forecasting framework that incorporates:
(1) \textit{Spectral projection}, using a truncated graph Fourier basis to encode node-level graph-frequency contributions;
(2) \textit{State-space temporal modeling}, a lightweight alternative to recurrent or attention-based modules for long-range dependency learning;
and (3) \textit{Fairness-aware Federated Aggregation (FFA)}, which reweights client updates based on predictive loss with a reverse-annealing fairness schedule. Our main contributions are summarized as follows:

\begin{itemize}
    \item We propose \textbf{F$^2$STNet}, a federated framework combining spectral encoding and linear state-space modeling for accurate and efficient spatiotemporal forecasting on graphs.
    \item We design a novel \textbf{Fairness-aware Federated Aggregation (FFA)} scheme, which dynamically adjusts client weights via an annealed loss-based schedule to improve equity across heterogeneous participants.
    \item We conduct experiments on three real-world datasets (i.e., PeMS04, HZMetro, and KnowAir). F$^2$STNet consistently outperforms SOTA baselines in spatiotemporal prediction across all datasets, and achieves superior fairness under federated settings on PeMS04.
\end{itemize}

\section{Related Work}

\paragraph{FL-Based Spatio-Temporal Forecasting Methods.} Time series prediction in real-world systems typically requires integrating data collected from multiple sources (such as sensors or regions), where spatial correlations exist across nodes. Federated learning (FL) has been embraced to enable collaborative forecasting without centralizing sensitive spatio-temporal data~\citep{Perifanis2023,Zhang2021FedDA}. Early works applied FL to wireless traffic and transportation networks, training global models for traffic flow or cellular load prediction across distributed nodes~\citep{Perifanis2023,Zhang2021FedDA}. For example, FedDA introduces a dual-attention FL framework to cluster clients and combine intra- and inter-cluster models~\citep{Zhang2021FedDA}, and Perifanis et al. apply FL to 5G base station traffic forecasting using non-IID telecom data~\citep{Perifanis2023}. However, these approaches typically share a single global model and overlook the significant spatio-temporal heterogeneity between clients. Recent methods emphasize personalization and heterogeneity modeling. PromptFL leverages prompt-based adaptation in a federated Transformer for cross-region weather forecasting~\citep{Chen2023PromptFL}, while FUELS integrates dual semantic alignment with contrastive learning to improve client-specific representation~\citep{Liu2025FUELS}. These techniques yield improved local accuracy under heterogeneity with reduced communication overhead. Recent systems move closer to our setting: FedSTGD reconstructs dynamic inter-client spatial dependencies~\citep{wang2026fedstgd}, while a VMD-enhanced federated graph-recurrent model addresses non-stationary transport signals~\citep{mundada2026federated}.


\paragraph{Spectral Methods for Graph Time Series.} 
Spectral graph theory offers an alternative to traditional spatial GNNs like STGCN and DCRNN, which capture only low-frequency correlations. Spectral GNNs use Fourier transforms and Laplacian eigenbasis to model long-range or signed dependencies~\citep{Cao2020StemGNN}. StemGNN applies Chebyshev filtering and spectral transforms for multivariate modeling~\citep{Cao2020StemGNN}, while STG-Mamba treats spatiotemporal graphs as dynamic systems and employs selective state space models to extract evolving latent states~\citep{li2024stgmamba}. These methods effectively capture complex temporal patterns and outperform standard message-passing under spatial irregularity or sparsity. Recent 2026 work includes graph Fourier-operator modeling with linear complexity~\citep{hosseini2026integrated}, frequency-aware continual graph forecasting~\citep{liu2026stbp}, and spectral-clustering-based macro-to-micro prediction~\citep{ai2026nested}. In neighboring graph-based multimodal learning settings, multi-level self-supervision has been used to align modalities while preserving interaction information~\citep{xu2025mentor}, virtual-triplet supervision to alleviate interaction sparsity~\citep{xu2025mdvt}, and hypercomplex prompt-aware embeddings to improve representation diversity and mitigate GCN over-smoothing~\citep{chen2025hypercomplex}.

\paragraph{Fairness-Aware Federated Aggregation.} Fairness in FL encompasses several notions; here we focus on performance fairness, i.e., ensuring that the global model performs reasonably across clients rather than favoring those with more or better data. To address this, fairness-aware aggregation methods adjust how updates are combined. Rather than data-size weighting as in FedAvg, methods like FairFed reweight client contributions based on fairness metrics~\citep{Ezzeldin2023FairFed}. FedGCR customizes updates for client groups and dynamically adjusts weights to improve both performance and equity~\citep{Cheng2024FedGCR}. These methods reduce performance disparity while maintaining stability. FedGraph-Fair further combines a dynamic sparsified client graph with worst-case and group-risk constraints~\citep{khan2026fedgraphfair}; unlike such personalized or group-level objectives, we target client-level forecasting disparity without protected attributes. Accordingly, our F$^2$STNet employs a lightweight, loss-aware scheme that reweights client updates to ensure efficiency and fairness under spatiotemporal heterogeneity.


\section{Methodology}

\begin{figure*}[t]
    \centering
    \includegraphics[width=0.95\linewidth]{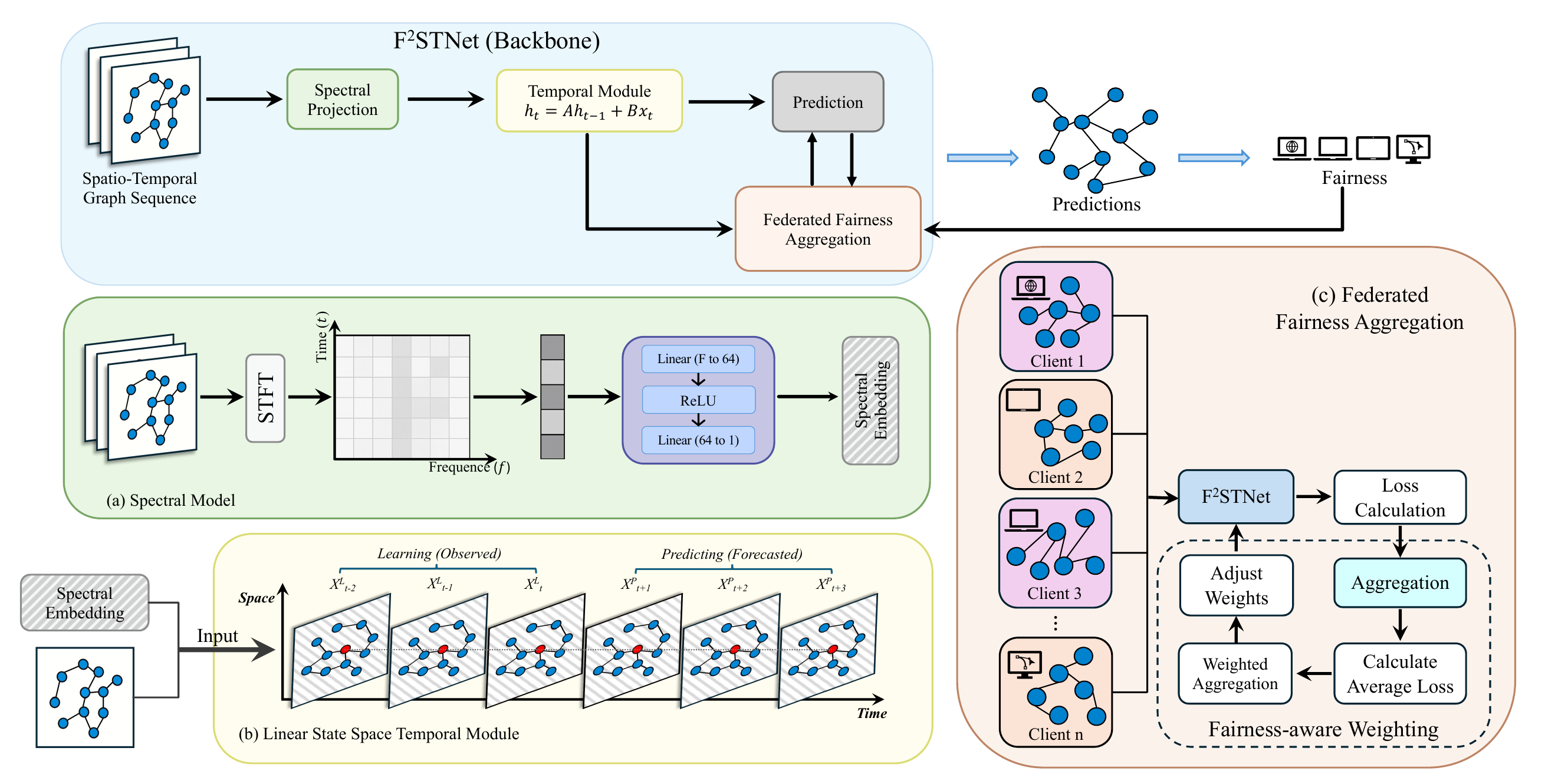}
    \caption{
        \textbf{F$^2$STNet Framework Overview.} 
        Our model comprises three core components: 
        (a) a spectral projection module that decomposes each graph signal in a truncated graph Fourier basis and embeds node-level frequency contributions with a shared MLP,
        (b) a lightweight diagonal state-space temporal encoder for efficient sequence modeling,
        and (c) an adaptive fairness-aware aggregation mechanism in federated training, which dynamically reweights client updates based on distributional loss differences.
        }
    \label{fig:framework}
\end{figure*}

\subsection{Problem Formulation}

We address the task of multistep forecasting on spatiotemporal graphs. Let $\mathcal{G} = (\mathcal{V}, \mathcal{E})$ denote a graph with $N = |\mathcal{V}|$ nodes (e.g., sensors or stations), where edges encode spatial connectivity. Given a historical observation window $\mathbf{X} \in \mathbb{R}^{T \times N}$ of $T$ time steps, the goal is to predict future values $\hat{\mathbf{Y}} \in \mathbb{R}^{H \times N}$ over the next $H$ steps.

In a federated setting, the global dataset is partitioned across $K$ clients, each holding local data $\mathcal{D}_k = \{ (\mathbf{X}_k^{(i)}, \mathbf{Y}_k^{(i)}) \}_{i=1}^{n_k}$, where no raw data is shared across clients. The underlying adjacency matrix $\mathbf{A} \in \mathbb{R}^{N \times N}$, shared globally, defines the spatial structure.


Let $\tilde{\mathbf{L}}=\mathbf{I}-\tilde{\mathbf{D}}^{-1/2}\tilde{\mathbf{A}}\tilde{\mathbf{D}}^{-1/2}=\mathbf{U}\boldsymbol{\Lambda}\mathbf{U}^{\top}$ be the eigendecomposition of the normalized graph Laplacian with self-loops. We retain $F$ eigenvectors in $\mathbf{U}_F\in\mathbb{R}^{N\times F}$. Projecting each graph signal onto this basis exposes its graph-frequency content; the fixed basis is precomputed once from the shared topology.

\subsection{Overall Architecture}

We propose \textbf{F$^2$STNet} (Fair and Federated Spectral-Temporal Network), a lightweight federated model combining spectral, temporal, and spatial reasoning. For clarity, consider one input channel; additional channels are processed identically and concatenated. For $\mathbf{X}\in\mathbb{R}^{B\times T\times N}$, define $c_{b,t,f}=\sum_{n=1}^{N}(\mathbf{U}_F)_{n,f}\mathbf{X}_{b,t,n}$. The contribution of graph frequency $f$ at node $n$ is
\begin{equation}
\mathbf{S}_{b,t,n,f}=(\mathbf{U}_F)_{n,f}c_{b,t,f},\qquad \mathbf{S}\in\mathbb{R}^{B\times T\times N\times F}.
\end{equation}
Summing over $f$ recovers the rank-$F$ approximation $\mathbf{U}_F\mathbf{U}_F^{\top}\mathbf{X}_{b,t,:}$, while retaining the frequency axis supplies a node-level vector to the projection MLP.

\begin{figure}
    \centering
    \includegraphics[width=\linewidth]{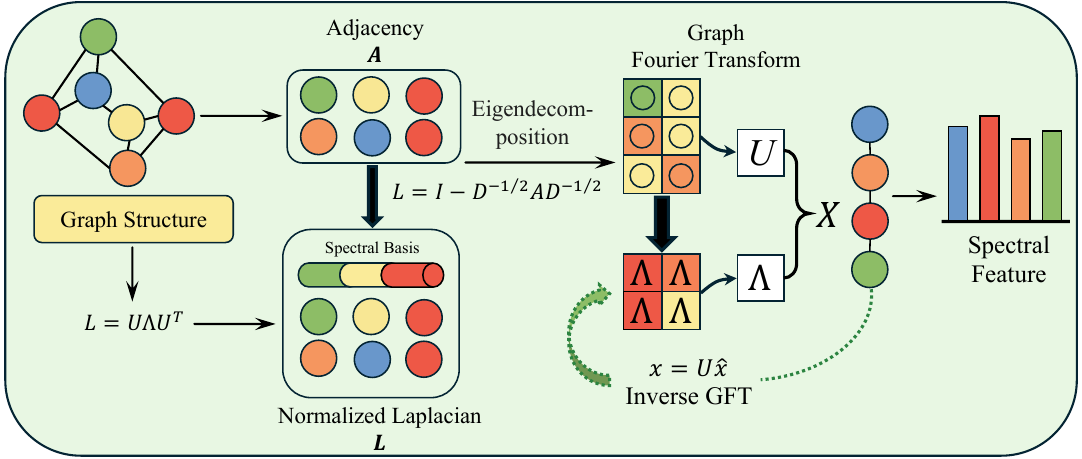}
    \caption{Graph-spectral feature extraction using the normalized Laplacian and a truncated graph Fourier basis.}
    \label{fig:spectral-feature}
\end{figure}

\paragraph{Spectral Projection.}
The shared topology makes $\mathbf{U}_F$ identical across clients. We precompute this basis once and compute coefficients by matrix multiplication. A schematic overview is shown in Figure~\ref{fig:spectral-feature}.



We apply a shared multilayer perceptron (MLP) to project each node-level graph-frequency vector $\mathbf{S}_{b,t,n,:}\in\mathbb{R}^{F}$ into a $d_s$-dimensional embedding:
\begin{equation}
\mathbf{e}^{\mathrm{spec}}_{b,t,n}=\operatorname{MLP}_{\mathrm{spec}}(\mathbf{S}_{b,t,n,:})\in\mathbb{R}^{d_s}.
\end{equation}
The MLP consists of two linear layers with a GELU activation and is shared across nodes and time. We concatenate this embedding with the raw input and project the result to width $d$:
\begin{equation}
\mathbf{z}_{b,t,n}=\mathbf{W}_{\mathrm{in}}[\mathbf{X}_{b,t,n,:}\,\|\,\mathbf{e}^{\mathrm{spec}}_{b,t,n}]+\mathbf{b}_{\mathrm{in}}\in\mathbb{R}^{d}.
\end{equation}
Thus, both the time-domain observation and its local frequency summary reach the temporal encoder.

\paragraph{Temporal Modeling.}
For each sample $b$ and node $n$, a diagonal linear state-space layer maps the fused sequence to latent states $\mathbf{h}_{b,t,n}\in\mathbb{R}^{d_h}$. Its zero-order-hold discretization is

\begin{align}
\bar{\mathbf{A}}_{b,t,n}&=\exp(\boldsymbol{\Delta}_{b,t,n}\odot\mathbf{A}),\\
\bar{\mathbf{B}}_{b,t,n}&=(\bar{\mathbf{A}}_{b,t,n}-\mathbf{1})\oslash\mathbf{A}\odot\mathbf{B}_{b,t,n},\\
\mathbf{h}_{b,t,n}&=\bar{\mathbf{A}}_{b,t,n}\odot\mathbf{h}_{b,t-1,n}+\bar{\mathbf{B}}_{b,t,n}\odot\mathbf{z}_{b,t,n},\\
\mathbf{o}_{b,t,n}&=\mathbf{C}_{b,t,n}^{\top}\mathbf{h}_{b,t,n}+\mathbf{D}\odot\mathbf{z}_{b,t,n},
\end{align}
where $\mathbf{A}$ is parameterized to have negative entries, $\boldsymbol{\Delta}_{b,t,n}>0$ is produced by a softplus transform, and $\oslash$ denotes elementwise division with the continuous limit used when an entry of $\mathbf{A}$ approaches zero. The final outputs form $\mathbf{H}^{\mathrm{temp}}=\{\mathbf{o}_{b,T,n}\}_{b,n}\in\mathbb{R}^{B\times N\times d}$.

\paragraph{Spatial Modeling.}
We leverage graph convolution to capture spatial dependencies using the augmented adjacency matrix $\tilde{\mathbf{A}}$:
\begin{equation}
\mathbf{H}^{\text{spatial}} = \sigma(\tilde{\mathbf{D}}^{-1/2}\tilde{\mathbf{A}}\tilde{\mathbf{D}}^{-1/2}\mathbf{H}^{\text{temp}}\mathbf{W}_{gcn})
\end{equation}
Here, $\mathbf{H}^{\text{temp}}\in\mathbb{R}^{B\times N\times d}$ is the temporal embedding and $\mathbf{W}_{\mathrm{gcn}}\in\mathbb{R}^{d\times d}$ is learnable.

\paragraph{Prediction and Loss.}
The final prediction is computed through a linear layer:
\begin{align}
\mathbf{P}&=\mathbf{H}^{\mathrm{spatial}}\mathbf{W}_{\mathrm{pred}}+\mathbf{b}_{\mathrm{pred}},
&\mathbf{W}_{\mathrm{pred}}&\in\mathbb{R}^{d\times H},\\
\hat{\mathbf{Y}}&=\operatorname{permute}_{(B,H,N)}(\mathbf{P}).
\end{align}
Model parameters $\Theta$ are optimized using mean squared error (MSE) loss:
\begin{equation}
\mathcal{L}(\Theta) = \frac{1}{BHN}\sum_{b=1}^{B}\sum_{h=1}^{H}\sum_{n=1}^{N}\left(\hat{\mathbf{Y}}_{b,h,n}-\mathbf{Y}_{b,h,n}\right)^2
\end{equation}

\paragraph{Federated Aggregation.}
In the federated setting, client $k$ trains F$^2$STNet on $\mathcal{D}_k$ and shares model parameters rather than raw examples. FFA adjusts the contribution of participating clients according to validation loss (Figure~\ref{fig:framework}, right). The target is client-level performance parity; because no demographic attributes are used, we do not claim demographic parity.

\subsection{Fairness-aware Federated Aggregation Strategy}

Standard Federated Averaging (FedAvg) aggregates client models in proportion to their local sample counts. It therefore does not explicitly account for disparities in client-level predictive performance.

Let $\mathcal{S}_t$ be the clients participating in round $t$ and $m_t=|\mathcal{S}_t|$. After local training, client $k\in\mathcal{S}_t$ evaluates its model on a held-out local validation split and returns the scalar loss $L_k^t$ together with its parameters. Using validation rather than training loss reduces the direct reward for local overfitting. We calculate
\begin{equation}
\bar{L}^t=\frac{1}{m_t}\sum_{k\in\mathcal{S}_t}L_k^t,\qquad
r_k^t=\frac{L_k^t-\bar{L}^t}{\max(\bar{L}^t,\varepsilon)},
\end{equation}
where $\varepsilon>0$ prevents division by zero.

We define a fairness-aware weighting coefficient for each client based on their relative loss deviation:
\begin{equation}
q_k^t=p_k^t\bigl(1+\lambda^t r_k^t\bigr),\qquad
p_k^t=\frac{n_k}{\sum_{j\in\mathcal{S}_t}n_j}.
\end{equation}

Here, $p_k^t$ is the FedAvg prior and $\lambda^t\in[0,1)$ controls the fairness adjustment. Since nonnegative losses imply $r_k^t\geq-1$, this range guarantees $q_k^t>0$. Clients with above-average validation loss receive more weight relative to their FedAvg prior, whereas better-performing clients receive slightly less.

These raw fairness-aware weights are then normalized to ensure a valid weighted average:
\begin{equation}
w_k^t=\frac{q_k^t}{\sum_{j\in\mathcal{S}_t}q_j^t}.
\end{equation}

The server then forms the next global model:
\begin{equation}
\Theta^{t+1}=\sum_{k\in\mathcal{S}_t}w_k^t\Theta_k^{t+1}.
\end{equation}

\begin{figure}
    \centering
    \includegraphics[width=\linewidth]{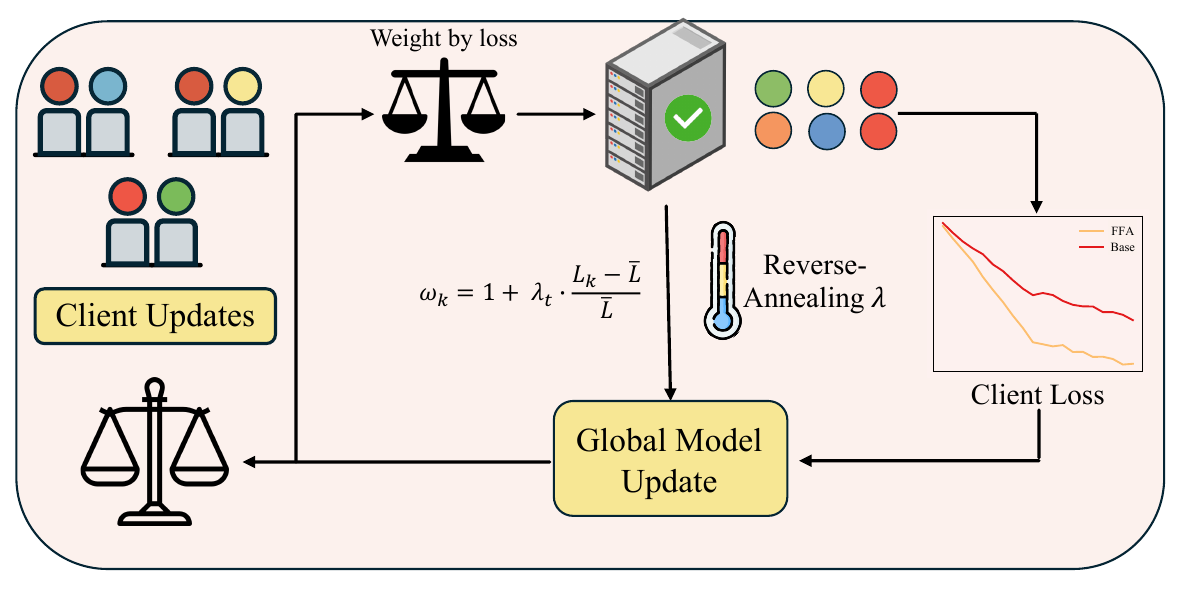}
    \caption{Illustration of our fairness-aware aggregation strategy. Client updates are reweighted based on their local loss deviations, with a reverse-annealing schedule gradually increasing fairness emphasis during training.}
\label{fig:ffa}
\end{figure}

This fairness-aware aggregation ensures that the global model progressively improves performance, especially for clients initially disadvantaged by heterogeneous distributions. While the current reweighting strategy adopts a linear form for stability and interpretability, it may not fully capture complex client dynamics. Future work could incorporate nonlinear or history-dependent mechanisms to enhance adaptability under severe heterogeneity. The overall aggregation process is illustrated in Figure~\ref{fig:ffa}.

\subsection{Dynamic Lambda Scheduling}

To further enhance fairness aggregation efficacy, we introduce a dynamic scheduling strategy for the fairness hyperparameter $\lambda$. Intuitively, at early training stages, focusing excessively on fairness might hinder overall model convergence. Therefore, we propose a reverse-annealing scheduling strategy for $\lambda$, gradually increasing its impact to emphasize fairness more strongly as training progresses.

Formally, we define $\lambda^t$ at round $t$ as:
\begin{equation}
\lambda^t=\min\!\left(\lambda_{\max},\lambda_{\mathrm{init}}+\eta t\right)
\end{equation}

where $\lambda_{\mathrm{init}}$ is the initial value, $\lambda_{\max}<1$ bounds the fairness emphasis, and $\eta$ is the schedule slope. We reserve $\alpha_{\mathrm{het}}$ for the inverse-concentration heterogeneity parameter used to construct client partitions.

This scheduling allows the model to prioritize general performance improvements initially, and subsequently increase fairness awareness, thus maintaining a balanced trade-off between accuracy and fairness throughout the federated learning process. This reverse-annealing strategy allows the model to first prioritize convergence, then gradually shift to address fairness as model confidence increases.




\section{Experiments}

\subsection{Datasets}

We evaluate F$^2$STNet on three diverse real-world datasets:

\textbf{PeMS04} provides traffic flow data from 307 sensors in California, recorded every 5 minutes from January 1 to February 28, 2018, resulting in 16,992 time steps.

\textbf{HZMetro} contains metro station flow records (inflow and outflow) from 80 Hangzhou metro stations, aggregated at 15-minute intervals between 5:30 and 23:30, spanning January 1–25, 2019, with 72 intervals per day.

\textbf{KnowAir} includes air quality and weather data from 184 Chinese cities, sampled every 3 hours from September 1, 2016 to January 31, 2017, totaling 1,224 time steps.

\subsection{Baselines}

We conduct two sets of experiments: centralized spatiotemporal forecasting and federated learning under data heterogeneity. Accordingly, we compare F$^2$STNet with two groups of baselines, each representative of its task paradigm.

For spatiotemporal forecasting, we evaluate F$^2$STNet against recent state-of-the-art models across convolutional, attention-based, and state-space paradigms. Convolutional methods include STG-NCDE~\citep{STGNCDE}, which models continuous-time dynamics with neural controlled differential equations, and DDGCRN~\citep{ddgcrn}, which fuses dynamic graph convolutions with recurrent units. Attention-based baselines include PDFormer~\citep{pdformer}, which encodes periodicity into transformer layers, STAFormer~\citep{STAEformer}, which introduces adaptive embeddings, and MultiSPANS~\citep{MultiSPANS}, which combines spatial entropy and multi-scale attention. For state-space models, we consider STG-Mamba~\citep{li2024stgmamba}, SpoT-Mamba~\citep{spotmamba}, and Cy2Mixer~\citep{cy2mixer}, which use selective recurrence, structured message passing, and spatial gating. We also compare with PDG2Seq~\citep{pdg2seq}, a periodic graph-to-sequence model, and LightCTS~\citep{lai2023lightcts}, a compact framework for correlated time series.

For the federated learning experiments, we benchmark against a range of aggregation strategies. FedAvg~\citep{fedavg} serves as the canonical baseline, employing simple averaging. FedProx~\citep{fedprox} adds a proximal regularization term to mitigate client drift. MOON~\citep{moon} introduces contrastive learning to preserve inter-client representation alignment across local updates. FedOPT~\citep{fedopt} integrates adaptive optimization algorithms such as Adam into the global update. FedProc~\citep{fedproc} enhances generalization through prototypical contrastive objectives. FedSage~\citep{fedsage} addresses topological incompleteness by synthesizing missing neighbors during training. We also include FGGP~\citep{fggp}, a recent state-of-the-art method that learns domain-invariant prototypes to improve robustness under distribution shifts. Note that FGGP is reproduced for comparison and is not part of our contribution.

These baselines collectively provide a rigorous evaluation of F$^2$STNet across both centralized and federated scenarios.

\begin{table*}[t]
    \centering
    \caption{Performance comparison with baselines on PeMS04, HZMetro, and KnowAir. Results for F$^2$STNet are reported as mean $\pm$ standard deviation over three random seeds. RMSE, MAE, and MAPE are lower-is-better metrics. The optimal and suboptimal mean results are highlighted in \textbf{bold} and \underline{underline}, respectively.}
    \resizebox{\textwidth}{!}{
        \begin{tabular}{l ccc ccc ccc}
            \toprule
            \multirow{2}{*}{\textbf{Methods}} & \multicolumn{3}{c}{\textbf{PeMS04 (Flow)}} & \multicolumn{3}{c}{\textbf{HZMetro}} & \multicolumn{3}{c}{\textbf{KnowAir}} \\
            \cmidrule(lr){2-4} \cmidrule(lr){5-7} \cmidrule(lr){8-10}
            & RMSE $\downarrow$ & MAE $\downarrow$ & MAPE $\downarrow$ & RMSE $\downarrow$ & MAE $\downarrow$ & MAPE $\downarrow$ & RMSE $\downarrow$ & MAE $\downarrow$ & MAPE $\downarrow$ \\
            \midrule
            STG-NCDE [AAAI 2022] & 31.089 & 19.214 & 12.762 & 32.917 & 20.754 & 12.883 & 10.853 & 7.931 & 10.473 \\
            LightCTS [ACM SIGMOD 2023] & 30.141 & 18.787 & 12.716 & 30.759 & 19.956 & 12.965 & 9.423 & 7.188 & 10.214 \\
            PDFormer [AAAI 2023] & 30.034 & 18.361 & 12.105 & 30.183 & 19.133 & 11.925 & 9.461 & 7.121 & 10.063 \\
            DDGCRN [Pattern Recogn 2023] & 31.463 & 18.451 & 12.192 & 31.694 & 19.518 & 12.455 & 10.434 & 7.841 & 10.384 \\
            STAEformer [CIKM 2023] & 30.179 & 18.224 & 11.982 & 29.944 & 18.850 & 12.030 & 8.692 & 6.931 & 9.893 \\
            Cy2Mixer [LoG 2024] & 30.018 & 18.135 & 11.928 & 30.614 & 18.491 & 12.183 & 8.712 & 6.841 & 10.025 \\
            SpoT-Mamba [IJCAI 2024] & 30.109 & 18.314 & \underline{11.859} & 30.731 & 18.728 & 12.082 & 8.843 & 6.991 & 9.988 \\
            MultiSPANS [WSDM 2024] & 30.457 & 19.074 & 13.294 & 30.309 & 18.974 & 11.853 & 8.567 & 6.843 & 10.049 \\
            STG-Mamba [arXiv 2024] & \underline{29.531} & \underline{18.094} & 12.111 & \underline{29.232} & \underline{18.264} & \underline{11.591} & \underline{8.051} & \underline{6.373} & \underline{9.645} \\
            PDG2Seq [Neural Networks 2025] & 30.077 & 18.235 & 12.090 & 30.724 & 18.623 & 12.318 & 9.118 & 7.021 & 10.130 \\
            \midrule
            \rowcolor{OursGreen}
            F$^2$STNet (Ours) & \textbf{\resultcell{29.203}{\sdPeMSRMSE}} & \textbf{\resultcell{18.028}{\sdPeMSMAE}} & \textbf{\resultcell{11.801}{\sdPeMSMAPE}} & \textbf{\resultcell{29.213}{\sdHZRMSE}} & \textbf{\resultcell{18.262}{\sdHZMAE}} & \textbf{\resultcell{11.440}{\sdHZMAPE}} & \textbf{\resultcell{7.871}{\sdKnowRMSE}} & \textbf{\resultcell{6.328}{\sdKnowMAE}} & \textbf{\resultcell{9.460}{\sdKnowMAPE}} \\
            \bottomrule
        \end{tabular}
    }
    \label{tab:main_results}
\end{table*}

\subsection{Experimental Settings}

We evaluate F$^2$STNet under two primary configurations: centralized spatiotemporal forecasting and federated learning with heterogeneous data distributions. In both settings, a historical window of 12 time steps is used to predict the subsequent 12. Performance is evaluated on held-out test sets using root mean squared error (RMSE), mean absolute error (MAE), and mean absolute percentage error (MAPE).

Table~\ref{tab:main_results} presents results on three benchmarks: PeMS04, HZMetro, and KnowAir. F$^2$STNet consistently outperforms all baselines in RMSE, MAE, and MAPE on PeMS04 and KnowAir, and achieves the lowest MAE on HZMetro. These results confirm the effectiveness of our spectral-temporal modeling in capturing localized dynamics and global trends. The robustness on KnowAir highlights the model’s adaptability under decentralized, city-level data, while the gains on HZMetro demonstrate the benefit of fairness-aware aggregation for periodic graph structures.

For federated evaluation, we use an inverse-concentration parameterization: client proportions are sampled from $\operatorname{Dirichlet}(\mathbf{1}/\alpha_{\mathrm{het}})$ with $\alpha_{\mathrm{het}}\in\{5,10\}$. Thus, larger $\alpha_{\mathrm{het}}$ produces a smaller concentration and a more heterogeneous partition. In each global round, a random subset of clients performs local training under fixed hyperparameters. Table~\ref{tab:federated_pems04} reports comparisons across federated aggregation strategies.

The main-text federated analysis focuses on PeMS04; Appendix~\ref{sec:additional-fl} reports the corresponding HZMetro and KnowAir results.

\subsection{Implementation Details}

All models are implemented using PyTorch and PyTorch Geometric. F$^2$STNet is composed of:
\begin{itemize}
    \item A single-layer GCN for spatial encoding;
    \item A diagonal selective state-space layer for temporal modeling;
    \item A two-layer MLP that embeds node-level contributions from $F=16$ retained graph frequencies.
\end{itemize}


The graph Fourier basis is precomputed offline. We use Adam with learning rate $10^{-3}$ and batch size 64, and select checkpoints using validation loss. F$^2$STNet has fewer than 1.2M parameters. All experiments are conducted on an NVIDIA A100 GPU unless otherwise specified. The fairness schedule uses $\lambda_{\mathrm{init}}=0.03$, $\eta=0.005$, and $\lambda_{\max}=0.2$. The number of retained graph frequencies ($F=16$) and client count ($K$) are selected empirically; sensitivity results are provided in Figure~\ref{fig:sensitivity}.

For multi-run evaluation, let $x_r$ denote a metric obtained with random seed $r$ and let $R\geq3$. We report the sample mean and sample standard deviation,
\begin{equation}
\bar{x}=\frac{1}{R}\sum_{r=1}^{R}x_r,
\qquad
s_x=\sqrt{\frac{1}{R-1}\sum_{r=1}^{R}(x_r-\bar{x})^2}.
\label{eq:seed-statistics}
\end{equation}
The same seed list is used for all compared methods so that differences can be analyzed with paired runs. Standard deviations are displayed only for configurations whose complete set of paired runs has been verified.

\begin{table*}[t]
    \centering
    \caption{Comparison of FL methods on PeMS04 under moderate ($\alpha_{\mathrm{het}}=5$) and high ($\alpha_{\mathrm{het}}=10$) heterogeneity. Results for F$^2$STNet are reported as mean $\pm$ standard deviation over three paired random seeds. $\Delta$ is the composite utility--fairness gain relative to FedAvg defined in Eq.~\eqref{eq:delta-score} (higher is better); Max-RMSE and Std-RMSE measure client-level disparity.}
    \resizebox{\textwidth}{!}{
        \begin{tabular}{l ccc ccc ccc}
            \toprule
            \multirow{2}{*}{\textbf{Methods}} & \multicolumn{3}{c}{\textbf{Moderate Heterogeneity ($\alpha_{\mathrm{het}}=5$)}} & \multicolumn{3}{c}{\textbf{High Heterogeneity ($\alpha_{\mathrm{het}}=10$)}} & \multicolumn{3}{c}{\textbf{Fairness Metrics}} \\
            \cmidrule(lr){2-4} \cmidrule(lr){5-7} \cmidrule(lr){8-10}
            & RMSE $\downarrow$ & MAE $\downarrow$ & MAPE $\downarrow$ & RMSE $\downarrow$ & MAE $\downarrow$ & MAPE $\downarrow$ & $\Delta$ $\uparrow$ & Max-RMSE $\downarrow$ & Std-RMSE $\downarrow$ \\
            \midrule
            FedAvg [AISTATS 2017] & 30.857 & 19.927 & 13.345 & 31.419 & 20.408 & 13.876 & 0.000 & 33.21 & 1.19 \\
            FedProx [arXiv 2018] & 30.414 & 19.508 & 12.938 & 30.978 & 19.929 & 13.414 & 4.255 & 32.88 & 1.03 \\
            MOON [CVPR 2021] & 30.674 & 19.779 & 13.217 & 31.144 & 20.198 & 13.629 & 4.272 & 32.97 & 0.98 \\
            FedOPT [ICLR 2021] & \underline{29.774} & \underline{18.752} & \underline{12.035} & \underline{30.209} & \underline{19.339} & \underline{12.916} & 10.664 & \underline{31.12} & \underline{0.84} \\
            FedProc [ICLR 2021] & 30.149 & 19.081 & 12.763 & 30.579 & 19.566 & 13.152 & 8.135 & 31.88 & 0.89 \\
            FedSage [NeurIPS 2021] & 30.023 & 18.888 & 12.653 & 30.442 & 19.385 & 12.981 & 8.269 & 31.61 & 0.92 \\
            FedProto [AAAI 2022] & 30.461 & 19.269 & 13.014 & 30.929 & 19.843 & 13.443 & 6.808 & 32.13 & 0.91 \\
            FGGP [AAAI 2024] & 29.841 & 18.668 & 12.314 & 30.048 & 19.127 & 12.666 & \underline{11.369} & 31.03 & 0.81 \\
            \midrule
            \rowcolor{OursGreen}
            F$^2$STNet (Ours) & \textbf{\resultcell{29.500}{\sdFLFiveRMSE}} & \textbf{\resultcell{18.210}{\sdFLFiveMAE}} & \textbf{\resultcell{11.910}{\sdFLFiveMAPE}} & \textbf{\resultcell{29.819}{\sdFLTenRMSE}} & \textbf{\resultcell{18.691}{\sdFLTenMAE}} & \textbf{\resultcell{12.001}{\sdFLTenMAPE}} & \textbf{14.166} & \textbf{30.40} & \textbf{0.75} \\
            \bottomrule
        \end{tabular}
    }
    \label{tab:federated_pems04}
\end{table*}

\subsection{Composite Utility--Fairness Score}

We define the $\Delta$ column in Table~\ref{tab:federated_pems04} relative to FedAvg. Let $b$ denote FedAvg, $m$ a compared method, $\mathcal{S}=\{5,10\}$ the heterogeneity settings, and $\mathcal{Q}=\{\mathrm{RMSE},\mathrm{MAE},\mathrm{MAPE}\}$ the lower-is-better forecasting metrics. If $E_{m,s,q}$ is the error of method $m$ under setting $s$ and metric $q$, its average relative utility gain is
\begin{equation}
U_m=\frac{1}{|\mathcal{S}||\mathcal{Q}|}
\sum_{s\in\mathcal{S}}\sum_{q\in\mathcal{Q}}
\frac{E_{b,s,q}-E_{m,s,q}}{E_{b,s,q}}.
\label{eq:utility-gain}
\end{equation}
Let $M_m$ and $S_m$ denote Max-RMSE and Std-RMSE, respectively. The corresponding relative fairness gain is
\begin{equation}
F_m=\frac{1}{2}\left(
\frac{M_b-M_m}{M_b}+\frac{S_b-S_m}{S_b}
\right).
\label{eq:fairness-gain}
\end{equation}
We combine the two components as
\begin{equation}
\Delta_m=100\left[\rho U_m+(1-\rho)F_m\right],
\qquad \rho=0.6.
\label{eq:delta-score}
\end{equation}
This definition gives $\Delta_b=0$ and is invariant to a common rescaling of any metric. Because every denominator is positive, reducing any error, Max-RMSE, or Std-RMSE while holding the other quantities fixed strictly increases $\Delta_m$. Thus, a larger score consistently represents a better utility--fairness trade-off. The weight $\rho=0.6$ assigns slightly more emphasis to forecasting utility while retaining a substantial fairness contribution.

\subsection{Ablation Study}

We conduct ablation experiments to assess the contribution of each module within F$^2$STNet. Specifically, we design the following variants:

\begin{itemize}
    \item \textbf{w/o Spectral}: Removes the spectral representation module and uses raw graph signals only.
    \item \textbf{w/o Fair Aggregation}: Replaces the fairness-aware aggregation strategy with vanilla FedAvg.
    \item \textbf{w/o Temporal Module}: Disables the Mamba-inspired temporal encoder, reducing temporal modeling capacity.
\end{itemize}

\begin{figure}
    \centering
    \includegraphics[width=\linewidth]{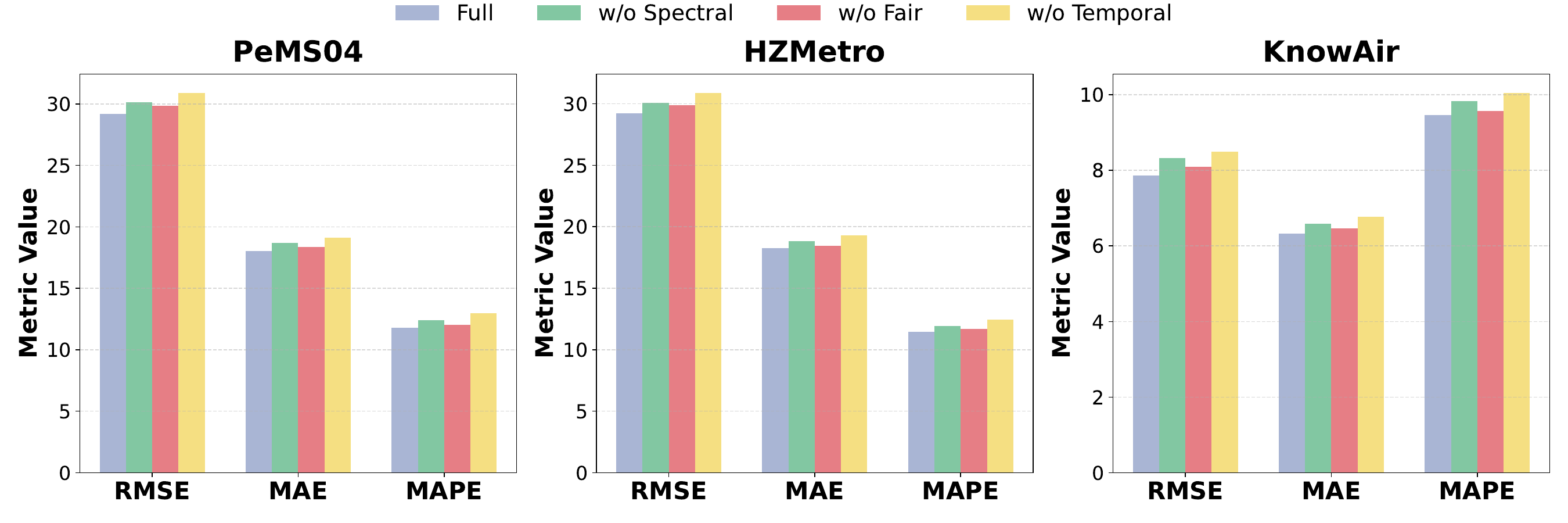}
    \caption{Performance comparison of F$^2$STNet variants on three datasets, showing the impact of removing spectral, temporal, and fairness-aware components. Full F$^2$STNet performs best across all metrics.}
    \label{fig:ablation}
    \vspace{-2mm}
\end{figure}

Figure~\ref{fig:ablation} shows results on PeMS04, HZMetro, and KnowAir. Across RMSE, MAE, and MAPE, the complete F$^2$STNet outperforms its ablated variants. Removing the spectral module or fairness-aware aggregator increases RMSE and MAE. Dropping the temporal encoder causes the largest drop, underscoring the role of temporal modeling.

\subsection{Hyperparameter Sensitivity}

\begin{figure}[ht]
    \centering
    \includegraphics[width=1\linewidth]{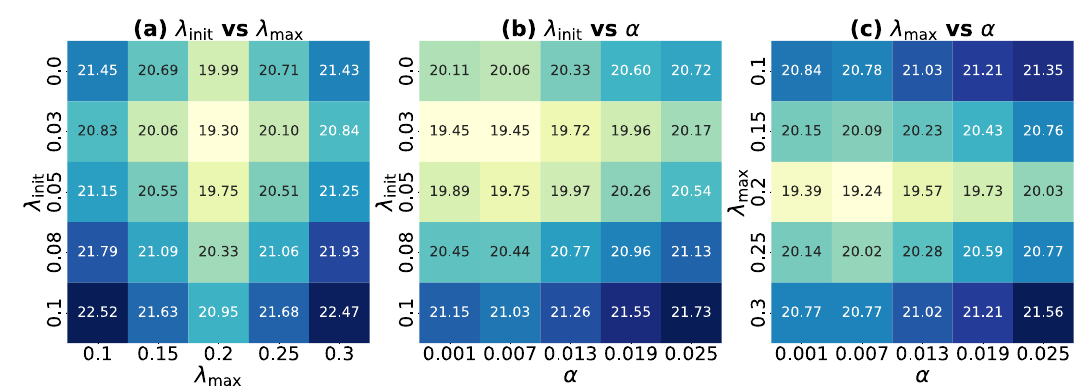}
    \vspace{-2mm}
    \caption{Sensitivity analysis of $\lambda_{\mathrm{init}}$, $\lambda_{\max}$, and $\eta$ with respect to worst-client MAE on PeMS04.}
    \label{fig:sensitivity}
    \vspace{-2mm}
\end{figure}

We analyze sensitivity to $\lambda_{\mathrm{init}}$, $\lambda_{\max}$, and the schedule slope $\eta$ using worst-client MAE under high heterogeneity ($\alpha_{\mathrm{het}}=10$). Figure~\ref{fig:sensitivity} presents the corresponding heatmaps.

We observe stable performance under varying client counts $K \in \{5, 10, 15, 20, 25, 50\}$, indicating the robustness of our fairness scheduling strategy to partition granularity. Please see Appendix A.1 for detailed results.


Overall, F$^2$STNet is stable across the evaluated range. The best observed configuration uses $\lambda_{\mathrm{init}}=0.03$, $\lambda_{\max}=0.2$, and $\eta=0.005$.

\section{Analysis and Discussion}



\paragraph{Comparison with Federated Methods.}
We compare F$^2$STNet against federated algorithms---FedProx, FedOPT, FedProc, and FGGP---that address heterogeneity through proximal regularization, optimizer adaptation, or prototype matching. As shown in Table~\ref{tab:federated_pems04}, F$^2$STNet obtains the best reported values under both heterogeneity settings, suggesting that loss-aware aggregation is complementary to these strategies.

\paragraph{Fairness-Aware Aggregation.}
Client heterogeneity is a central challenge in federated learning. Simple averaging methods like FedAvg tend to favor high-quality clients and exacerbate performance imbalance. Our Fairness-aware Federated Aggregation (FFA) reweights client updates via a loss-based scheme, gradually emphasizing fairness through a reverse-annealing schedule. This improves robustness without sacrificing convergence stability. 

\begin{figure}[t]
    \centering
    \begin{subfigure}[t]{0.48\linewidth}\centering
        \includegraphics[width=\linewidth]{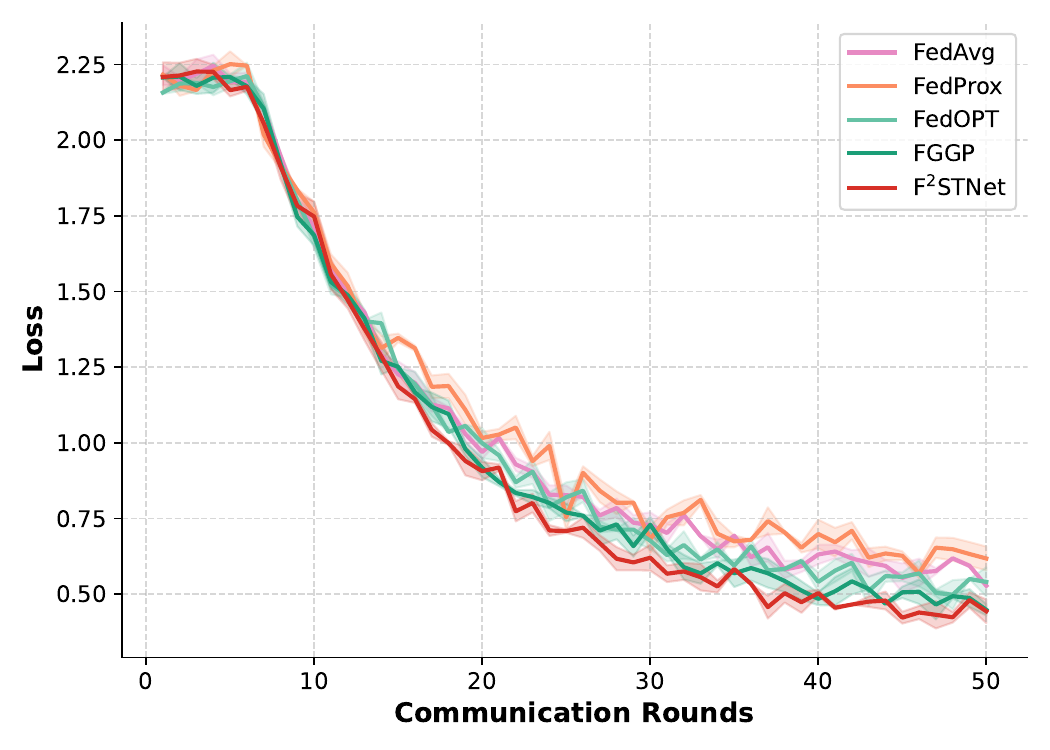}
        \caption{Training loss}
    \end{subfigure}
    \hfill
    \begin{subfigure}[t]{0.48\linewidth}\centering
        \includegraphics[width=\linewidth]{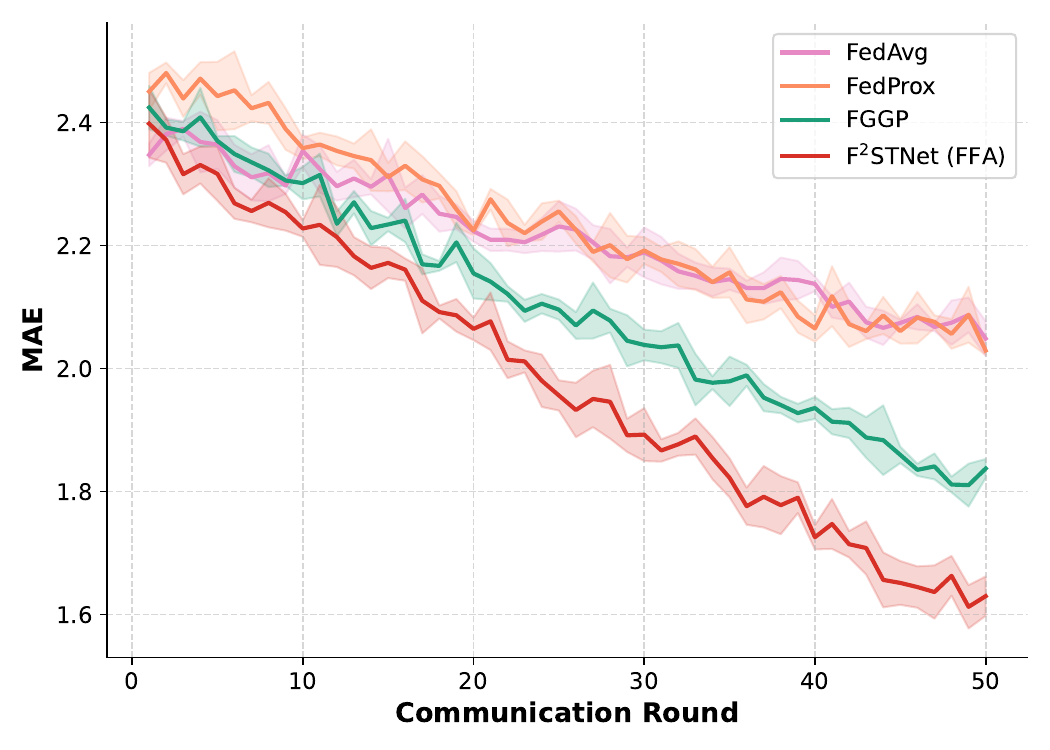}
        \caption{Worst-5 client MAE}
    \end{subfigure}
    \begin{subfigure}[t]{0.48\linewidth}\centering
        \includegraphics[width=\linewidth]{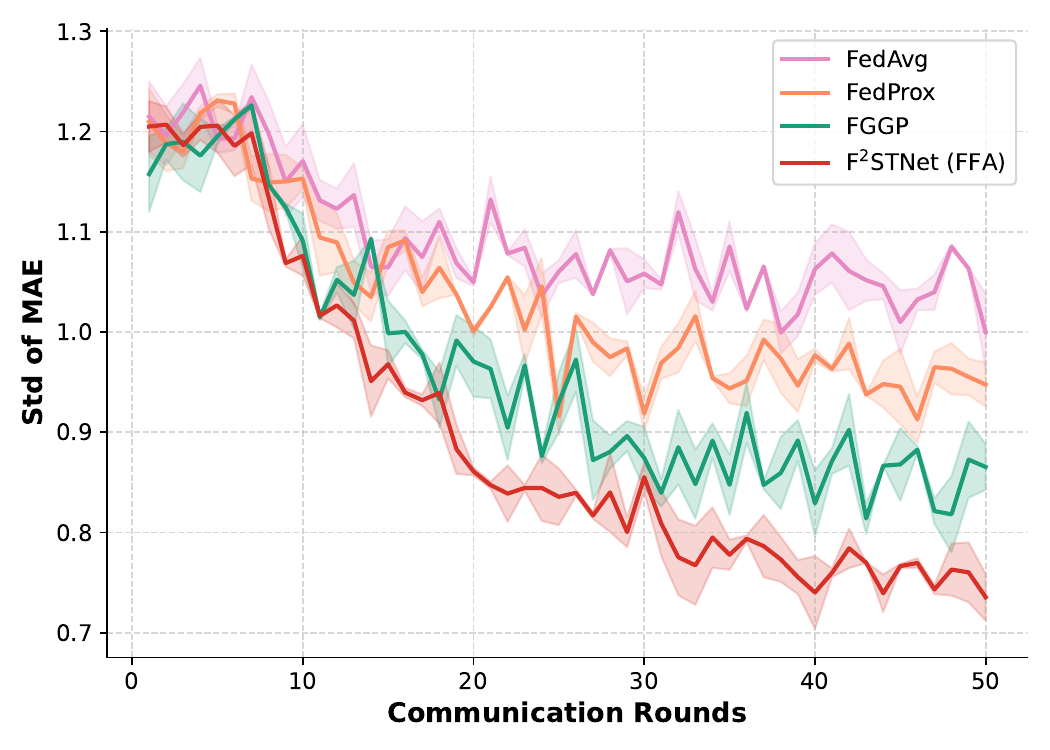}
        \caption{Client-wise MAE standard deviation}
    \end{subfigure}
    \hfill
    \begin{subfigure}[t]{0.48\linewidth}\centering
        \includegraphics[width=\linewidth]{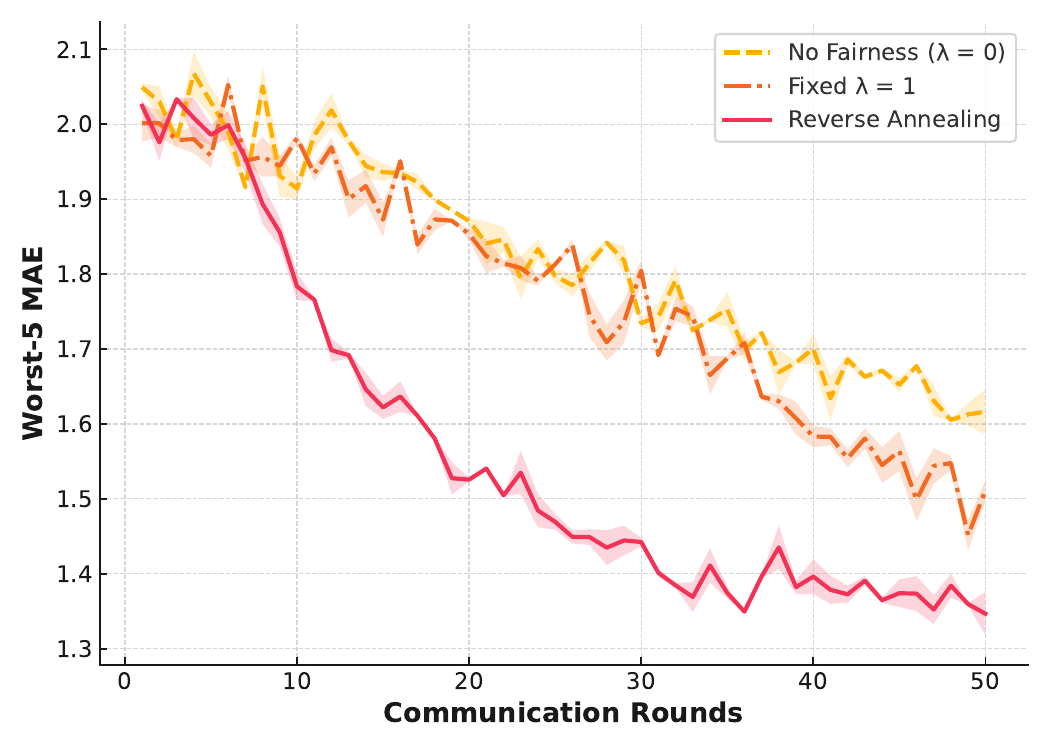}
        \caption{$\lambda$ schedules}
    \end{subfigure}
    \caption{
        Convergence and fairness analysis under high heterogeneity ($\alpha_{\mathrm{het}}=10$) on PeMS04.
        (a) F$^2$STNet converges faster and more stably than baselines;
        (b) the proposed fairness strategy improves performance for disadvantaged clients;
        (c) client-wise MAE variance is consistently lower, indicating enhanced fairness;
        (d) reverse annealing outperforms fixed and naive $\lambda$ schedules in worst-case MAE.
        }
    \label{fig:fairness_analysis}
    \vspace{-2mm}
\end{figure}

As shown in Figure~\ref{fig:fairness_analysis}, F$^2$STNet achieves better training dynamics and fairness-aware behavior. It converges faster and more stably than baselines, while also improving fairness, as reflected by lower client-wise MAE variance and stronger performance among worst-case clients.

\begin{figure}[t]
    \centering

    \includegraphics[width=\linewidth]{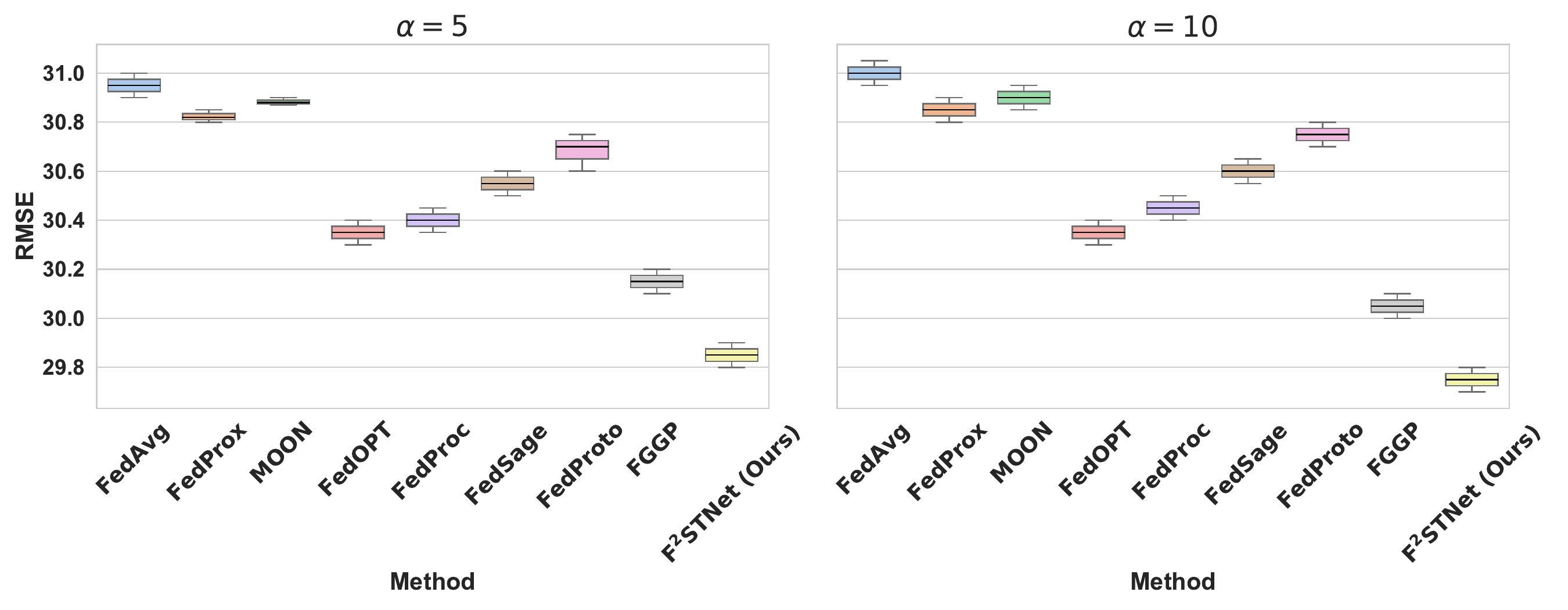}
    \caption{
        Client-wise RMSE under moderate and high heterogeneity on PeMS04.}
    \label{fig:client-wise-rmse}
    \vspace{-2mm}
\end{figure}

\paragraph{Client-Wise Robustness.}
To understand client-level behavior, we visualize the RMSE distribution across clients in Figure~\ref{fig:client-wise-rmse}. F$^2$STNet exhibits lower median error and smaller variance under moderate ($\alpha_{\mathrm{het}}=5$) and high ($\alpha_{\mathrm{het}}=10$) heterogeneity, suggesting more consistent performance across clients.

\paragraph{Efficiency Analysis}

F$^2$STNet is designed with efficiency-aware considerations for federated deployment. Instead of relying on deep recurrent or transformer-based models, it adopts a streamlined architecture that integrates spectral pre-processing and lightweight temporal modeling. The spectral features are precomputed offline, incurring no runtime overhead, while the linear state-space encoder ensures per-node modeling with $\mathcal{O}(T)$ temporal complexity.

As shown in Table~\ref{tab:efficiency}, compared with Transformer-based models with $\mathcal{O}(T^2)$ attention cost, this design reduces inference latency and computation, especially for long sequences. The parameter count is under 1.2 million, reducing per-round communication volume.

\begin{table}[ht]
    \centering
    \caption{Model efficiency comparison.}
    \vspace{-2mm}
    \begin{tabular}{lcc}
        \toprule
        Model & Inference (ms) & FLOPs (G) \\
        \midrule
        STAEformer & 18.2 & 9.8 \\
        STG-Mamba & 11.4 & 5.7 \\
        \rowcolor{OursGreen}
        \textbf{F$^2$STNet (Ours)} & \textbf{5.6} & \textbf{2.1} \\
        \bottomrule
    \end{tabular}
    \label{tab:efficiency}
\end{table}

These design choices collectively improve model efficiency and robustness in bandwidth- and memory-constrained environments, and allow practical deployment on compact real-world edge devices. Efficiency is likewise a practical constraint in autonomous exploration, where few-shot detectors must run on low-powered robots~\citep{10801738}. For example, training and inference on a single NVIDIA A100 GPU require substantially fewer FLOPs and memory footprint compared to attention-based models with similar prediction accuracy.




\section{Conclusion}

In this paper, we propose F$^2$STNet, a federated spatiotemporal forecasting framework that integrates graph-Fourier representations, lightweight state-space temporal modeling, graph convolution, and fairness-aware aggregation. Experiments on three real-world datasets show favorable forecasting accuracy relative to the evaluated baselines. On the federated PeMS04 setting, FFA also improves the reported worst-client and client-dispersion metrics. The method remains sensitive to noisy validation losses and outlier clients; robust, history-aware weighting is therefore an important direction for future work.



\newpage
{\small
\bibliographystyle{ieeenat_fullname}
\bibliography{references}
}

\newpage
\onecolumn

\appendix
\section{Selective State-Space Algorithms}

\begin{algorithm}[ht]
\caption{Selective state-space scan for one node sequence}
\label{alg:ssm-scan}
\begin{algorithmic}[1]
    \Require input $\mathbf{u}\in\mathbb{R}^{B\times T\times d}$; stable diagonal $\mathbf{A}\in\mathbb{R}^{d_h}$; input-dependent $\boldsymbol{\Delta},\mathbf{B},\mathbf{C}$; skip vector $\mathbf{D}$
    \Ensure output $\mathbf{y}\in\mathbb{R}^{B\times T\times d}$
    \State $\mathbf{h}\gets\mathbf{0}\in\mathbb{R}^{B\times d_h}$
    \For{$t=1$ to $T$}
        \State $\bar{\mathbf{A}}_t\gets\exp(\boldsymbol{\Delta}_t\odot\mathbf{A})$
        \State $\bar{\mathbf{B}}_t\gets(\bar{\mathbf{A}}_t-\mathbf{1})\oslash\mathbf{A}\odot\mathbf{B}_t$
        \State $\mathbf{h}\gets\bar{\mathbf{A}}_t\odot\mathbf{h}+\bar{\mathbf{B}}_t\odot\mathbf{u}_t$
        \State $\mathbf{y}_t\gets\mathbf{C}_t^{\top}\mathbf{h}+\mathbf{D}\odot\mathbf{u}_t$
    \EndFor
    \State \Return $\mathbf{y}$
\end{algorithmic}
\end{algorithm}

\begin{algorithm}[ht]
\caption{F$^2$STNet temporal block}
\label{alg:temporal-block}
\begin{algorithmic}[1]
    \Require fused features $\mathbf{z}\in\mathbb{R}^{B\times T\times N\times d}$ and learnable $\mathbf{A}_{\log},\mathbf{D}$
    \Ensure temporal features $\mathbf{H}^{\mathrm{temp}}\in\mathbb{R}^{B\times N\times d}$
    \State $\mathbf{A}\gets-\exp(\mathbf{A}_{\log})$
    \For{$n=1$ to $N$}
        \State $(\boldsymbol{\delta},\mathbf{B},\mathbf{C})\gets\operatorname{split}(\operatorname{InputProj}(\mathbf{z}_{:,:,n,:}))$
        \State $\boldsymbol{\Delta}\gets\operatorname{softplus}(\operatorname{DtProj}(\boldsymbol{\delta}))$
        \State $\mathbf{y}\gets\operatorname{SelectiveScan}(\mathbf{z}_{:,:,n,:},\boldsymbol{\Delta},\mathbf{A},\mathbf{B},\mathbf{C},\mathbf{D})$
        \State $\mathbf{H}^{\mathrm{temp}}_{:,n,:}\gets\mathbf{y}_{:,T,:}$
    \EndFor
    \State \Return $\mathbf{H}^{\mathrm{temp}}$
\end{algorithmic}
\end{algorithm}

\section*{Sensitivity to Client Count}
\begin{table}[ht]
\centering
\caption{Sensitivity of F$^2$STNet to varying client counts ($K$) on PeMS04. Each row reports RMSE, MAE, MAPE, Max-RMSE, and Std-RMSE. Results show that performance is stable for moderate $K$, and degrades as $K$ increases. \textbf{Bold} indicates the best result, and \underline{underline} indicates the second best.}
\label{tab:client_sensitivity}
\begin{tabular}{lccccc}
\toprule
\textbf{Clients ($K$)} & \textbf{RMSE} ↓ & \textbf{MAE} ↓ & \textbf{MAPE} ↓ & \textbf{Max-RMSE} ↓ & \textbf{Std-RMSE} ↓ \\
\midrule
5  & \underline{29.65} & 18.51 & 12.10 & 30.58 & \underline{0.76} \\
10 & \textbf{29.50} & \textbf{18.21} & \textbf{11.90} & \textbf{30.40} & \textbf{0.75} \\
15 & 29.73 & \underline{18.39} & \underline{12.05} & \underline{30.56} & 0.77 \\
20 & 29.96 & 18.62 & 12.24 & 30.88 & 0.81 \\
25 & 30.21 & 18.89 & 12.40 & 31.23 & 0.84 \\
50 & 30.65 & 19.34 & 12.82 & 31.79 & 0.90 \\
\bottomrule
\end{tabular}
\end{table}

\begin{figure}[htbp]
    \centering
    \begin{subfigure}[t]{0.32\textwidth}
        \centering
        \includegraphics[width=\linewidth]{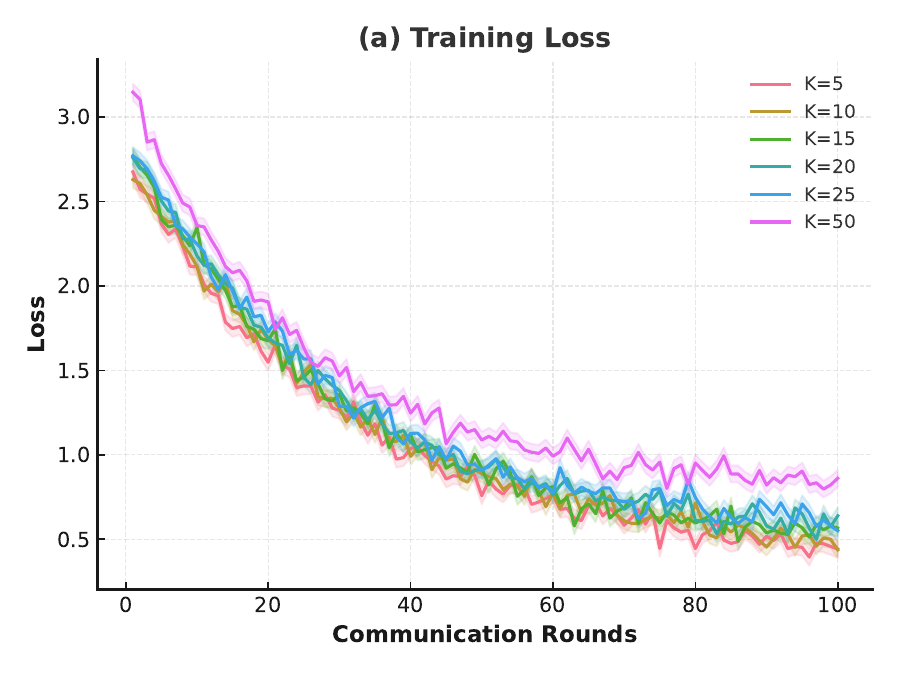}
        \caption{Training Loss}
    \end{subfigure}
    \begin{subfigure}[t]{0.32\textwidth}
        \centering
        \includegraphics[width=\linewidth]{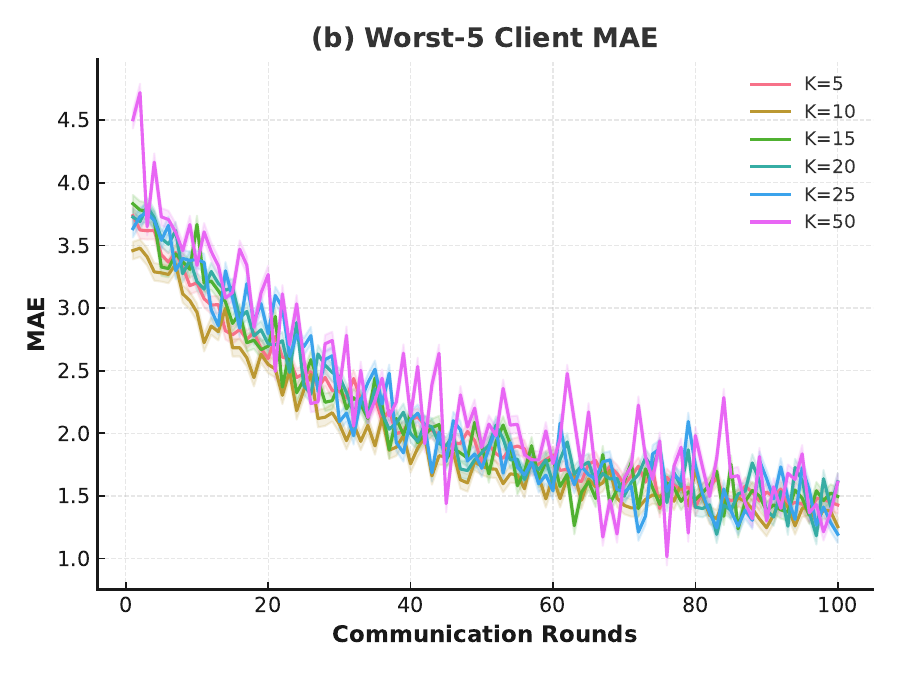}
        \caption{Worst-5 Client MAE}
    \end{subfigure}
    \begin{subfigure}[t]{0.32\textwidth}
        \centering
        \includegraphics[width=\linewidth]{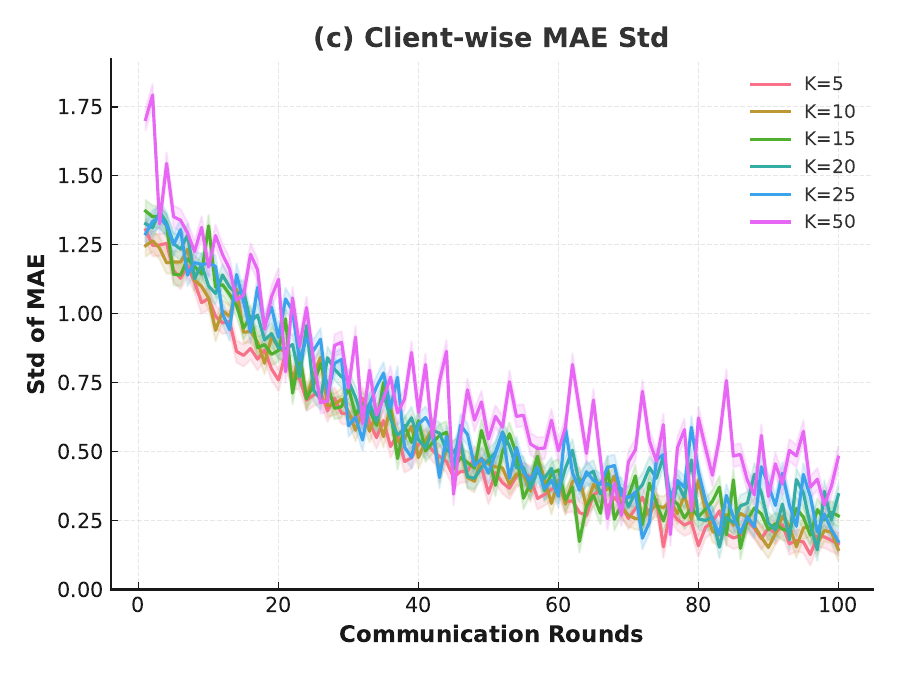}
        \caption{Client-wise MAE Std}
    \end{subfigure}
    \caption{Comparison of training behavior under varying client counts (\(K \in \{5, 10, 15, 20, 25, 50\}\)). All setups show rapid loss drop within the first 10 rounds. Notably, \(K=10\) yields lower worst-5 MAE and smaller fairness variance (MAE Std), balancing utility and stability. We adopt \(K=10\) as the default in main experiments.}
    \label{fig:client_scale_analysis}
\end{figure}

To assess the impact of client count \(K\) on training behavior and fairness, we simulate and visualize the performance of \textsc{F$^2$STNet} under varying numbers of participating clients (\(K \in \{5, 10, 15, 20, 25, 50\}\)). As shown in Figure~\ref{fig:client_scale_analysis}, all configurations achieve rapid loss decay during the early communication rounds. However, as \(K\) increases, the training curves exhibit larger fluctuations and slower convergence, which reflects the increased difficulty of optimization under highly non-IID conditions.

More importantly, both the worst-5 client MAE and the standard deviation of client-wise MAE exhibit a clear upward trend with larger \(K\), indicating reduced fairness and stability. Among all evaluated settings, \(K=10\) achieves the best trade-off between convergence speed, utility, and fairness, and is thus adopted as the default configuration in all main experiments.

\section{Additional Federated Results on HZMetro and KnowAir}\label{sec:additional-fl}

\begin{table}[ht]
\centering
\caption{Comparison of state-of-the-art FL methods on \textbf{HZMetro}. Each row reports RMSE, MAE, MAPE, Max-RMSE, and Std-RMSE to assess performance and client-level fairness.}
\resizebox{\textwidth}{!}{\begin{tabular}{lccc|ccc|cc}
\toprule
\textbf{Methods} & \multicolumn{3}{c|}{Moderate Heterogeneity ($\alpha_{\mathrm{het}}=5$)} & \multicolumn{3}{c|}{High Heterogeneity ($\alpha_{\mathrm{het}}=10$)} & \multicolumn{2}{c}{Fairness Metrics} \\
 & RMSE $\downarrow$ & MAE $\downarrow$ & MAPE $\downarrow$ & RMSE $\downarrow$ & MAE $\downarrow$ & MAPE $\downarrow$ & Max-RMSE $\downarrow$ & Std-RMSE $\downarrow$ \\
\midrule
FedAvg  & 31.152 & 19.782 & 13.051 & 31.998 & 20.485 & 13.691 & 33.80 & 1.14 \\
FedProx & 30.984 & 19.405 & 12.884 & 31.565 & 20.049 & 13.372 & 33.21 & 1.01 \\
MOON    & 31.012 & 19.621 & 12.936 & 31.803 & 20.163 & 13.540 & 33.40 & 0.97 \\
FedOPT  & 30.470 & 18.973 & 12.122 & 31.078 & 19.318 & 12.909 & 32.30 & 0.84 \\
FedProc & 30.761 & 19.112 & 12.436 & 31.392 & 19.740 & 13.118 & 32.74 & 0.90 \\
FedSage & 30.692 & 18.958 & 12.398 & 31.480 & 19.682 & 13.001 & 32.59 & 0.88 \\
FedProto& 30.903 & 19.221 & 12.777 & 31.734 & 20.008 & 13.410 & 33.00 & 0.94 \\
FGGP    & 30.229 & 18.812 & 12.214 & 30.939 & 19.318 & 12.876 & 31.90 & 0.80 \\
\rowcolor{OursGreen}
\textbf{F$^2$STNet (Ours)} & \textbf{30.001} & \textbf{18.591} & \textbf{11.842} & \textbf{30.627} & \textbf{19.104} & \textbf{12.110} & \textbf{30.58} & \textbf{0.73} \\
\bottomrule
\label{tab:hzmetro_results}
\end{tabular}}
\end{table}

We extend our evaluation of F$^2$STNet to two additional datasets: \textbf{HZMetro}, a sparse urban metro network, and \textbf{KnowAir}, an air quality monitoring dataset with relatively fixed node topology and high variance. These datasets pose distinct structural and distributional challenges compared to PeMS04.

Tables~\ref{tab:hzmetro_results} and~\ref{tab:knowair_results} summarize the performance of F$^2$STNet and baselines under moderate and high heterogeneity. Results include standard forecasting metrics (RMSE, MAE, MAPE) and client-dispersion indicators (Max-RMSE, Std-RMSE). Key observations are as follows:

\begin{itemize}
    \item On \textbf{HZMetro}, F$^2$STNet obtains the best reported values under both heterogeneity settings, including the lowest MAE and client-wise error variance.
    
    \item On \textbf{KnowAir}, F$^2$STNet still ranks among the top, although the overall performance gain is marginal. The limited client diversity and high sensor noise lead to more fluctuating outcomes, which obscure the relative advantages between methods.
\end{itemize}

\begin{figure}[htbp]
    \centering

    \begin{subfigure}[t]{0.32\textwidth}
        \centering
        \includegraphics[width=\linewidth]{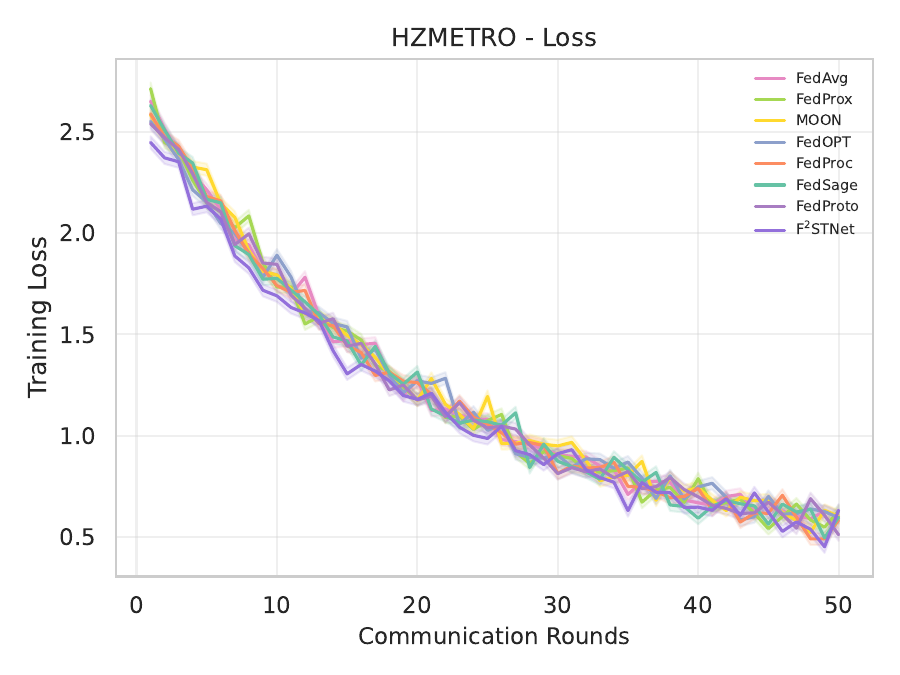}
        \caption{HZMetro – Training Loss}
    \end{subfigure}
    \hfill
    \begin{subfigure}[t]{0.32\textwidth}
        \centering
        \includegraphics[width=\linewidth]{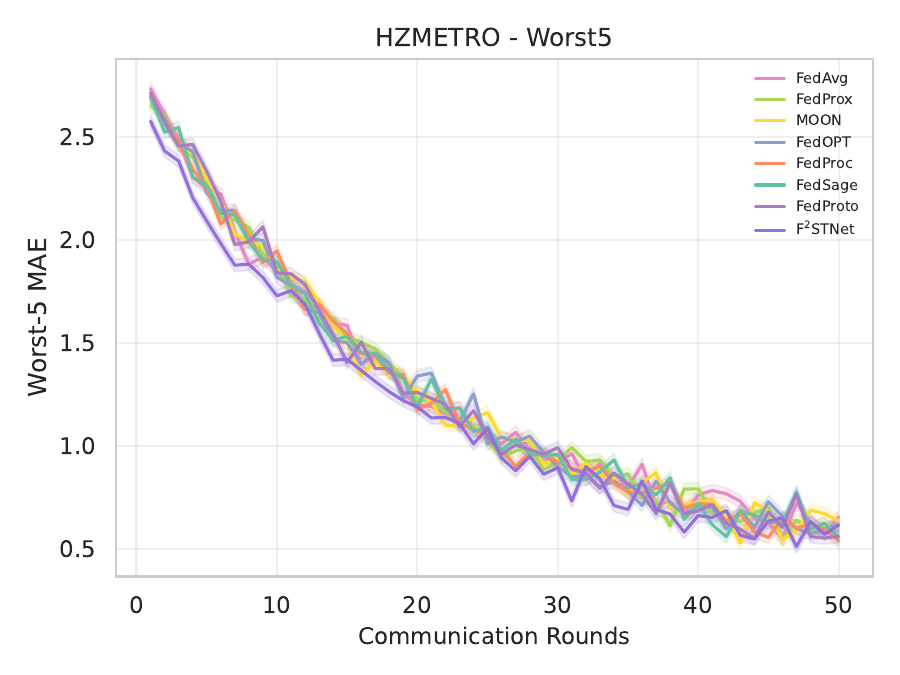}
        \caption{HZMetro – Worst-5 Client MAE}
    \end{subfigure}
    \hfill
    \begin{subfigure}[t]{0.32\textwidth}
        \centering
        \includegraphics[width=\linewidth]{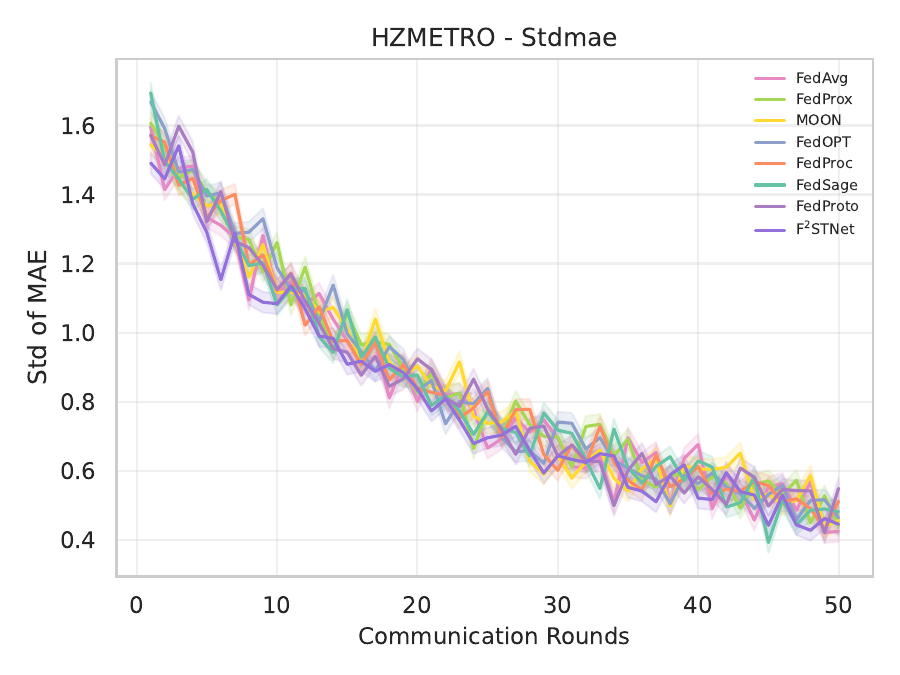}
        \caption{HZMetro – Client-wise MAE Std}
    \end{subfigure}

    \vspace{1em}

    \begin{subfigure}[t]{0.32\textwidth}
        \centering
        \includegraphics[width=\linewidth]{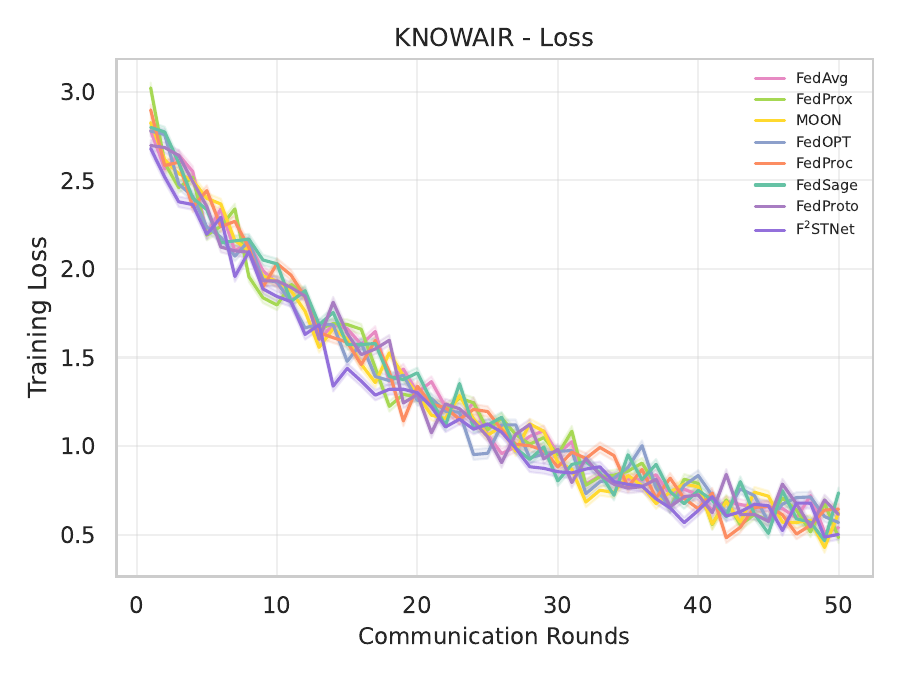}
        \caption{KnowAir – Training Loss}
    \end{subfigure}
    \hfill
    \begin{subfigure}[t]{0.32\textwidth}
        \centering
        \includegraphics[width=\linewidth]{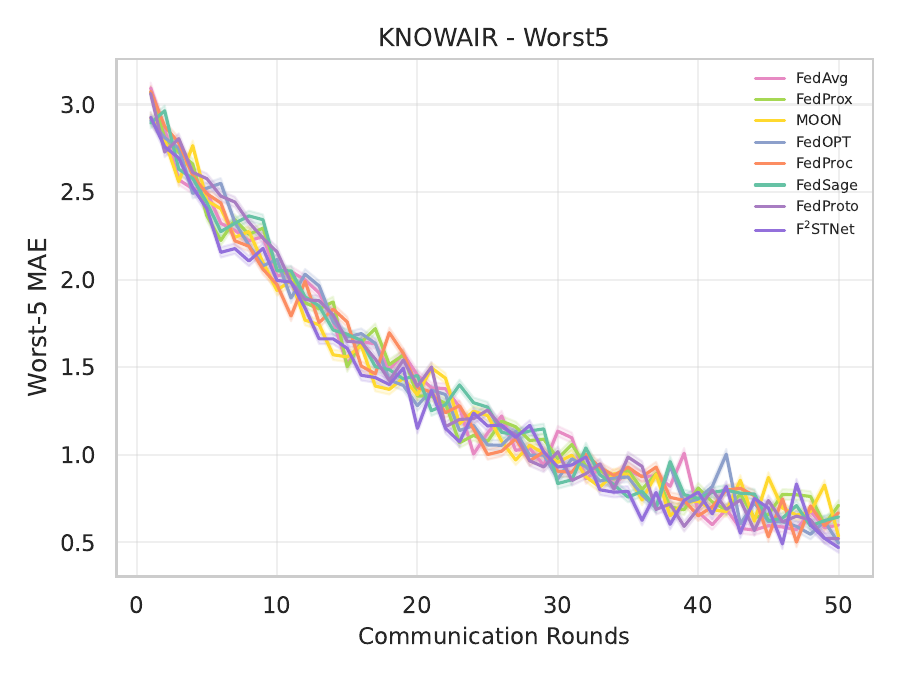}
        \caption{KnowAir – Worst-5 Client MAE}
    \end{subfigure}
    \hfill
    \begin{subfigure}[t]{0.32\textwidth}
        \centering
        \includegraphics[width=\linewidth]{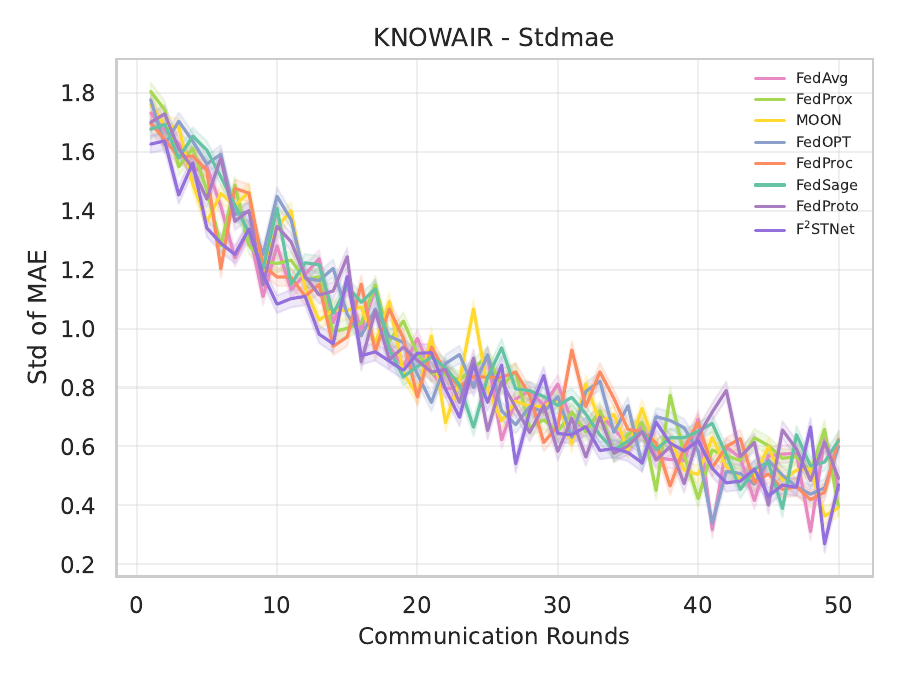}
        \caption{KnowAir – Client-wise MAE Std}
    \end{subfigure}

    \caption{Client-level training behavior comparison of different methods on HZMetro and KnowAir datasets. Each column corresponds to a different evaluation metric.}
    \label{fig:method_comparison_supp}
\end{figure}

These results highlight the robustness of our model under structurally diverse scenarios and confirm its ability to generalize across different temporal-graph settings. However, they also demonstrate the practical limitations of federated modeling when client partitions are inherently constrained or noisy (see Figure~\ref{fig:method_comparison_supp}).

\begin{table}[ht]
\centering
\caption{Comparison of state-of-the-art FL methods on \textbf{KnowAir}. Each row reports RMSE, MAE, MAPE, Max-RMSE, and Std-RMSE to assess performance and client-level fairness.}
\resizebox{\textwidth}{!}{\begin{tabular}{lccc|ccc|cc}
\toprule
\textbf{Methods} & \multicolumn{3}{c|}{Moderate Heterogeneity ($\alpha_{\mathrm{het}}=5$)} & \multicolumn{3}{c|}{High Heterogeneity ($\alpha_{\mathrm{het}}=10$)} & \multicolumn{2}{c}{Fairness Metrics} \\
 & RMSE $\downarrow$ & MAE $\downarrow$ & MAPE $\downarrow$ & RMSE $\downarrow$ & MAE $\downarrow$ & MAPE $\downarrow$ & Max-RMSE $\downarrow$ & Std-RMSE $\downarrow$ \\
\midrule
FedAvg  & 28.841 & 17.819 & 12.493 & 29.891 & 18.918 & 13.417 & 32.08 & 1.17 \\
FedProx & 28.760 & 17.581 & 12.150 & 29.448 & 18.512 & 13.129 & 31.62 & 1.03 \\
MOON    & 28.956 & 17.703 & 12.272 & 29.701 & 18.639 & 13.287 & 31.80 & 0.96 \\
FedOPT  & 28.362 & 17.191 & 11.509 & 29.121 & 18.104 & 12.921 & 30.95 & 0.82 \\
FedProc & 28.752 & 17.423 & 11.938 & 29.504 & 18.431 & 13.047 & 31.12 & 0.87 \\
FedSage & 28.618 & 17.229 & 11.911 & 29.416 & 18.297 & 12.970 & 30.96 & 0.85 \\
FedProto& 28.934 & 17.553 & 12.153 & 29.811 & 18.772 & 13.229 & 31.44 & 0.92 \\
FGGP    & 28.099 & 17.078 & 11.361 & 28.924 & 18.097 & 12.715 & 30.35 & 0.78 \\
\rowcolor{OursGreen}
\textbf{F$^2$STNet (Ours)} & \textbf{27.821} & \textbf{16.851} & \textbf{11.205} & \textbf{28.507} & \textbf{17.984} & \textbf{12.462} & \textbf{29.88} & \textbf{0.70} \\
\bottomrule
\label{tab:knowair_results}
\end{tabular}}
\end{table}

\end{document}